\pdfoutput=1
\documentclass[cameraready]{Interspeech}

\title{ProSarc: Prosody-Aware Sarcasm Recognition Framework via Temporal Prosodic Incongruity}

\author[affiliation={1}, orcid=0009-0000-2164-310X]{Prathamjyot}{Singh}
\author[affiliation={2}, orcid=0000-0002-7014-2210]{Ashima}{Sood}
\author[affiliation={3}, orcid=0000-0002-3187-4929, correspondingauthor]{Sahil}{Sharma}
\author[affiliation={1}, orcid=0000-0003-2585-2827]{Jasmeet}{Singh}

\address{
    $^1$ Department of Computer Science and Engineering, Thapar Institute of Engineering and Technology, Patiala, India \\
    $^2$ School of Computing, Engineering and Intelligent Systems, Ulster University, Londonderry, United Kingdom \\
    $^3$ School of Computing, Ulster University, Belfast, United Kingdom
}

\email{psingh1\_be22@thapar.edu, sood-a1@ulster.ac.uk, s.sharma@ulster.ac.uk, jasmeet.singh@thapar.edu}

\keywords{Sarcasm Detection, Speech Prosody, Temporal Prosodic Incongruity, Uncertainty Estimation}

\usepackage{comment}
\usepackage{float}
\usepackage{longtable}

\begin{document}

\maketitle

\begin{abstract}


    We present ProSarc, an audio-only framework that detects sarcasm by modelling \emph{temporal prosodic incongruity}, that is, the mismatch between local prosodic dynamics and the utterance-level emotional baseline.  Dual encoding paths, a Global Emotion Encoder and a Temporal Prosody Encoder (BiLSTM + multi-head attention), feed a Prosodic Incongruity Analyzer that produces a scalar incongruity score for classification.  Monte Carlo dropout provides uncertainty estimates, and an attention-based mechanism localises sarcastic onset without frame-level labels.  ProSarc outperforms prior audio-only methods on MUStARD++ (F1\,=\,75.3) and generalises to spontaneous (PodSarc, F1\,=\,62.9) and cross-lingual speech (MuSaG, F1\,=\,65.6).  Ten-run validation confirms the contribution of incongruity modelling (Wilcoxon $p{=}0.002$, Cohen's $d{=}1.51$).  Human evaluation shows that model uncertainty tracks perceptual ambiguity and predicted onsets align with human-annotated temporal windows.
\end{abstract}

\section{Introduction}
\label{sec:intro}
 
Sarcasm in spoken language is conveyed largely through prosodic cues, including pitch contour, timing, and intensity, rather than through lexical markers alone~\cite{Rockwell2000,cheang2009,Bryant11102010}. While multimodal systems that fuse text, audio, and vision have advanced sarcasm detection on benchmarks such as MUStARD++~\cite{ray-etal-2022-multimodal,castro-etal-2019-towards}, audio is typically treated as an auxiliary signal, and detection still relies heavily on textual or visual cues~\cite{poria-etal-2019-meld,schuller2018}.
 
Audio-only approaches have received comparatively less attention. Existing methods predominantly operate on utterance-level acoustic statistics such as mean pitch, global energy, and spectral summaries that discard the temporal dynamics through which sarcasm is actually realised~\cite{ray-etal-2022-multimodal,castro-etal-2019-towards,gao22}. Recurrent and attention-based architectures implicitly encode some temporal structure~\cite{vaswani2023attentionneed}, yet no prior audio-only system explicitly represents sarcasm as a \emph{measurable incongruity} between fine-grained local prosodic dynamics and the utterance-level emotional baseline.
 
We address this gap with \textbf{ProSarc} (Prosody-Aware Sarcasm Recognition), an audio-only framework that models sarcasm as \emph{temporal prosodic incongruity}.  The model encodes audio through two parallel paths, a Global Emotion Encoder for utterance-level prosodic statistics and a Temporal Prosody Encoder for frame-level dynamics, and fuses them via a Prosodic Incongruity Analyzer that produces an explicit scalar incongruity score for classification.  Monte Carlo dropout~\cite{uncertainity} provides uncertainty estimates, and an attention-based mechanism offers weak temporal localisation of sarcastic onset without requiring frame-level annotations.
 
Our contributions are as follows:
\begin{enumerate}[leftmargin=*,nosep]
  \item We propose an audio-only sarcasm detection framework that
    \emph{explicitly} models temporal prosodic incongruity between
    local dynamics and a global emotional baseline, rather than
    relying on implicit temporal encoding or the utterance-level
    statistics.
  \item We introduce a weak-supervision mechanism for temporal
    sarcasm onset estimation using attention-weighted per-frame
    divergence, providing an interpretable temporal analysis without
    requiring fine-grained annotations.
  \item We evaluate on four benchmarks spanning scripted,
    spontaneous, and
    cross-lingual settings,
    demonstrating consistent improvements over prior audio-only
    methods with rigorous statistical validation (Wilcoxon
    $p{=}0.002$, Cohen's $d{=}1.51$).
    \item We incorporate MC~dropout uncertainty estimation and
    validate through human evaluation that model confidence
    tracks perceptual ambiguity: predicted uncertainty
    identifies a zone of genuine inter-annotator disagreement
    ($\kappa{=}0.34$), and a monotonic confidence gradient
    aligns with human consensus.
\end{enumerate}
 
\section{Related Work}
 
\noindent\textbf{Prosodic Foundations of Sarcasm} 
Psycholinguistic evidence establishes that sarcasm is signalled through characteristic prosodic patterns. Rockwell~\cite{Rockwell2000} showed that sarcastic speech exhibits lower pitch, slower rate, and greater intensity relative to sincere utterances.  Cheang and Pell~\cite{cheang2009} extended this finding cross-linguistically, identifying language-specific pitch and timing cues in both Cantonese and English.  Bryant~\cite{Bryant11102010} further demonstrated that prosodic exaggeration and contrastive stress serve as key markers of irony.  More recently, F\"{u}nfgeld et al.~\cite{funfgeld25_interspeech} confirmed that German listeners rely on prosodic cues for irony perception, and Li et al.~\cite{li2024functionaltradeoffprosodicsemantic} showed a functional trade-off between prosodic and semantic channels in conveying sarcasm.  Together, these studies motivate modelling sarcasm as a prosodic phenomenon rather than a purely lexical one.
 
\noindent\textbf{Computational Sarcasm Detection} 
Early computational work relied on textual features such as lexical incongruity and sentiment polarity~\cite{joshi17,ghosh-veale-2016-fracking,khodak-etal-2018-large}. Multimodal datasets including MUStARD~\cite{castro-etal-2019-towards} and MUStARD++~\cite{ray-etal-2022-multimodal} enabled fusion-based approaches across text, audio, and vision, with methods such as the Multimodal Transformer~\cite{TsaiBLKMS19} improving performance through cross-modal attention.  However, most multimodal systems are dominated by textual or visual cues and do not explicitly model fine-grained prosodic dynamics~\cite{Gao2025SpokenIJ}.
 
Audio-only sarcasm detection has received less attention. Gao et al.~\cite{gao22} applied CNN-based transfer learning to speech, while Raghuvanshi et al.~\cite{Raghuvanshi2025IntramodalRA} showed that intra-modal temporal inconsistencies are a hallmark of sarcastic intent.  Recent work has focused on scalability through LLM-aided annotation for large-scale podcast data~\cite{li2025leveraginglargelanguagemodels} and cross-lingual evaluation on German~\cite{scott2025musagmultimodalgermansarcasm}. Nevertheless, existing audio-only systems rely on utterance-level features or implicit temporal encoding~\cite{adieu16}, without explicitly representing sarcasm as a measurable incongruity between local prosodic dynamics and global emotional context.
 
\noindent\textbf{Uncertainty in Speech and NLP} 
Predictive uncertainty is increasingly valued for robustness under ambiguity.  Gal and Ghahramani~\cite{uncertainity} introduced MC dropout as a practical Bayesian approximation, while Lakshminarayanan et al.~\cite{lakshminarayanan2017simplescalablepredictiveuncertainty} showed that deep ensembles achieve stronger calibration.  Kendall and Gal~\cite{kendall} decomposed predictive uncertainty into aleatoric and epistemic components for vision tasks.  Despite its utility, uncertainty estimation has seen limited adoption in sarcasm detection.  ProSarc incorporates MC dropout to provide an interpretable confidence signal without altering classification decisions.

\begin{figure*}[t]
  \centering
  \includegraphics[width=\linewidth, trim=0 90pt 150pt 0, clip]{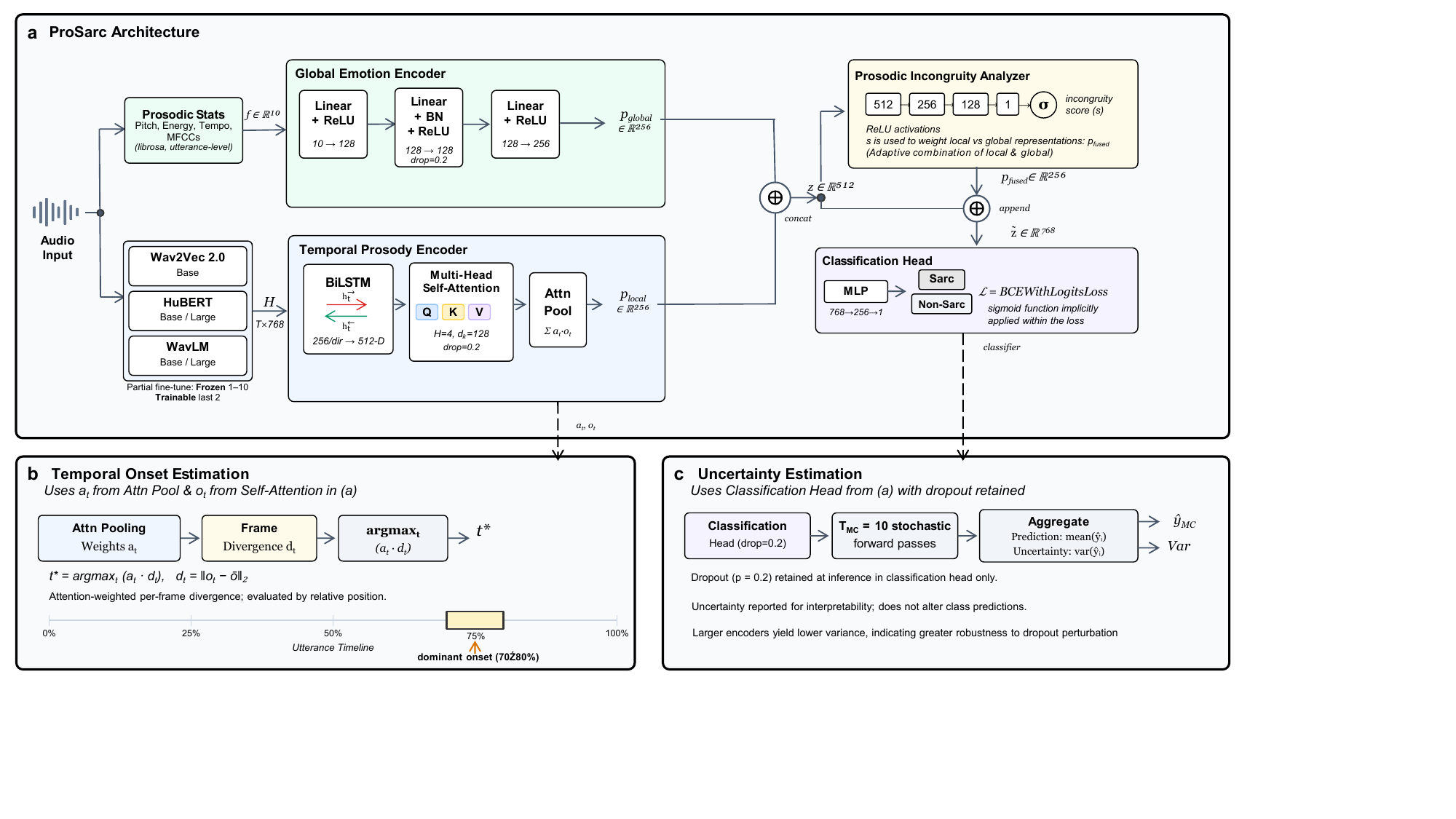}
  \caption{Architecture of ProSarc.
    \textbf{(a)}~Audio is processed via two parallel paths:
    the \emph{Global Emotion Encoder}
    (librosa utterance-level prosodic statistics $\to$ 3-layer MLP
    with dropout $\to$
    $\mathbf{p}_{\text{global}} \in \mathbb{R}^{256}$) and the
    \emph{Temporal Prosody Encoder}
    (SSL encoder $\to$ BiLSTM $\to$ multi-head self-attention $\to$
    attention pooling $\to$
    $\mathbf{p}_{\text{local}} \in \mathbb{R}^{256}$).
    The concatenated $\mathbf{z} \in \mathbb{R}^{512}$ is scored
    by the \emph{Prosodic Incongruity Analyzer}
    (MLP $\to$ sigmoid $\to$ $s$), fused to form $\mathbf{h} \in \mathbb{R}^{768}$,
    and classified via a single logit with weighted BCE loss.
    \textbf{(b)}~Temporal sarcasm onset estimated from
    attention-weighted per-frame divergence $a_t \cdot d_t$
    (dashed arrow: reuses $a_t$, $\mathbf{o}_t$ from~(a)).
    \textbf{(c)}~Predictive uncertainty via MC~dropout
    ($T_{\text{MC}}{=}10$, $p{=}0.2$; dashed arrow: reuses
    classifier from~(a)).}
  \label{fig:prosodydecoder}
\end{figure*}

\section{Methodology}
\label{sec:proposed_approach}

Given an audio utterance $\mathbf{A}$ with label $y\in\{0,1\}$, ProSarc predicts sarcasm from audio alone.  Our main hypothesis is that sarcasm appears in the form of \emph{temporal prosodic incongruity} which refers to utterance-wise prosodic trends that differ from the computed emotional baseline.  As illustrated in Figure~\ref{fig:prosodydecoder}, the model encodes audio through two parallel paths: a \emph{Global Emotion Encoder} that captures utterance-level statistics, and a \emph{Temporal Prosody Encoder} that models frame-level dynamics. Their outputs are fused by a \emph{Prosodic Incongruity Analyzer} that produces an explicit scalar incongruity score, which is then used for final classification.

\subsection{Dual-Path Prosodic Encoding}
\label{ssec:encoding}

\noindent\textbf{Global Emotion Encoder}
For each utterance we extract a 10-dimensional prosodic feature vector consisting of pitch statistics (mean, standard deviation, minimum, and maximum), energy statistics (mean and standard deviation), speaking rate estimated via the zero-crossing rate, spectral features including spectral centroid and bandwidth, and the mean of the first Mel-frequency cepstral coefficient (MFCC$_1$).  All features are computed at the frame level (25\,ms window, 10\,ms hop) using \texttt{librosa} with 40~Mel filterbanks and aggregated to utterance-level statistics.

The resulting vector $\mathbf{f} \in \mathbb{R}^{10}$ is passed through a three-layer MLP:
\begin{equation}
  \mathbf{g} = \mathrm{MLP}(\mathbf{f}):\;
  10 \xrightarrow{\mathrm{ReLU}} 128
    \xrightarrow{\mathrm{BN,\,drop,\,ReLU}} 128
    \xrightarrow{\mathrm{ReLU}} 256
\end{equation}
with batch normalisation and dropout ($p{=}0.2$) at the second layer.  The final layer directly produces the global embedding $\mathbf{p}_{\text{global}} \in \mathbb{R}^{256}$, representing the expected emotional baseline of the utterance.

\noindent\textbf{Temporal Prosody Encoder}
A single pretrained self-supervised encoder Wav2Vec\,2.0 \cite{baevski2020wav2vec20frameworkselfsupervised}, HuBERT \cite{HsuBTLSM21}, or WavLM \cite{WavLM} is evaluated per experiment while the downstream architecture is held fixed.  We partially fine-tune each encoder by training only the last two transformer layers; lower layers are frozen as they encode general acoustic structure, whereas upper layers better capture task-relevant prosodic variation~\cite{dai21_interspeech}.  The encoder produces frame-level embeddings $\mathbf{H} \in \mathbb{R}^{T \times 768}$.

A double-layer bidirectional LSTM (hidden size 256 per direction) models temporal dependencies, producing forward ($\mathbf{h}_t^{\rightarrow}$) and backward ($\mathbf{h}_t^{\leftarrow}$) hidden states that are concatenated:
\begin{equation}
  \mathbf{h}_t =
    [\mathbf{h}_t^{\rightarrow};\mathbf{h}_t^{\leftarrow}]
    \in \mathbb{R}^{512}
\end{equation}


\begin{table*}[t]
\centering
\caption{Computational cost and resource usage of audio encoders with partial fine-tuning (last two layers unfrozen). CO\textsubscript{2} estimated per single training run assuming 30 epochs, NVIDIA Tesla T4 (70\,W TDP), and 0.385\,kg\,CO\textsubscript{2}/kWh (IEA 2022 global average).}
\label{tab:model_params}
\renewcommand{\arraystretch}{1.15}
\scriptsize
\begin{tabular}{|p{2.4cm}|p{1.8cm}|p{1.6cm}|p{2.0cm}|p{2.0cm}|p{1.8cm}|}
\hline
\textbf{Encoder} 
& \textbf{Trainable Params} 
& \textbf{Train Time / Epoch} 
& \textbf{Inference Time (MC=10)} 
& \textbf{GPU Memory (Train)} 
& \textbf{Est.\ CO\textsubscript{2} (g)\,$\downarrow$} \\
\hline
OpenSMILE + SVM 
& 6.37K features 
& 0.12 min 
& 0.08 s 
& CPU only 
& $<$\,1 \\
\hline
Wav2Vec2 Base 
& 7.9M 
& 1.65 min 
& 15.9 s 
& 6.9 GB 
& 22.2 \\
\hline
HuBERT Base 
& 8.1M 
& 1.71 min 
& 16.4 s 
& 7.0 GB 
& 23.0 \\
\hline
WavLM Base 
& 8.0M 
& 1.70 min 
& 16.9 s 
& 7.0 GB 
& 22.9 \\
\hline
HuBERT Large 
& 25.4M 
& 3.55 min 
& 33.5 s 
& 12.4 GB 
& 47.8 \\
\hline
WavLM Large 
& 25.2M 
& 3.42 min  
& 31.6 s 
& 11.8 GB 
& 46.1 \\
\hline
\end{tabular}
\end{table*}

A multi-head self-attention layer ($H{=}4$, $d_k{=}128$, dropout
$p{=}0.2$) captures salient prosodic:
\begin{equation}
  \mathbf{O} = \mathrm{softmax}\Bigl(
    \frac{\mathbf{Q}\mathbf{K}^{\top}}{\sqrt{d_k}}
  \Bigr)\mathbf{V}
\end{equation}
where $\mathbf{Q}, \mathbf{K}, \mathbf{V}$ are linear projections
of the BiLSTM output sequence and $\mathbf{o}_t \in \mathbb{R}^{512}$
denotes the attention output at frame~$t$.
The attention outputs are projected to a compact local embedding
using a linear layer:
\begin{equation}
  \tilde{\mathbf{o}}_t = \mathbf{W}_p \mathbf{o}_t + \mathbf{b}_p,
  \quad \mathbf{W}_p \in \mathbb{R}^{256 \times 512},
  \quad \tilde{\mathbf{o}}_t \in \mathbb{R}^{256}
\end{equation}
Temporal aggregation uses learned attention-weighted pooling with
a projection vector $\mathbf{w} \in \mathbb{R}^{256}$:
\begin{equation}
  a_t = \frac{\exp(\mathbf{w}^{\top} \tilde{\mathbf{o}}_t)}
             {\sum_{j=1}^{T} \exp(\mathbf{w}^{\top} \tilde{\mathbf{o}}_j)},
  \quad
  \mathbf{p}_{\text{local}} = \sum_{t=1}^{T} a_t \, \tilde{\mathbf{o}}_t
  \in \mathbb{R}^{256}
\end{equation}
where $a_t$ are scalar attention weights summing to one.

\subsection{Incongruity Modeling and Classification}
\label{ssec:incongruity}

The two path outputs are concatenated as
$\mathbf{z} = [\mathbf{p}_{\text{local}}; \mathbf{p}_{\text{global}}] \in \mathbb{R}^{512}$
and passed through a dedicated MLP followed by a sigmoid to produce an explicit scalar \textit{incongruity score} $s$:
\[
s = \sigma\big(\mathrm{MLP}(\mathbf{z})\big): 512 \to 256 \to 128 \to 1
\]
where $\sigma(\cdot)$ denotes the sigmoid function, ensuring $s \in [0,1]$ and intermediate layers use ReLU activation. No auxiliary supervision is applied to~$s$; it is learned directly as a gating signal and used to perform adaptive fusion between the local and global representations.

The predicted incongruity score is then used to perform
adaptive fusion between the local and global representations.
Specifically, $s$ controls the relative contribution of global prosodic
information, while $(1-s)$ weights the local temporal representation:
\begin{equation}
\mathbf{p}_{\text{fused}} = (1-s)\,\mathbf{p}_{\text{local}} + s\,\mathbf{p}_{\text{global}}.
\end{equation}
\(\mathbf{p}_{\text{fused}} \in \mathbb{R}^{256}\), since it is a weighted sum of two 256-dimensional vectors. The fused representation is concatenated with the original combined
embedding to preserve complementary information:
\begin{equation}
\mathbf{h} = [\mathbf{p}_{\text{fused}};\mathbf{z}] \in \mathbb{R}^{768},
\end{equation}
and projected through a fusion MLP to obtain a final utterance-level representation. This design ensures that both the adaptively fused and the original embeddings contribute to the final representation, enhancing stability and interpretability.

The model is trained end-to-end using weighted binary
cross-entropy with logits:
\begin{equation}
\mathcal{L}
= -\bigl[
y \log \sigma(\hat{z})
+ (1-y)\log\bigl(1-\sigma(\hat{z})\bigr)
\bigr],
\end{equation}
where $\sigma(\cdot)$ denotes the sigmoid function implicitly applied within the loss, and class imbalance is addressed using positive class weighting.
All components such as pretrained encoder, both encoding paths,
incongruity analyzer, and classifier are optimized jointly under this single loss.

\subsection{Temporal Onset Estimation}
\label{ssec:onset}

We estimate when sarcastic cues emerge within an utterance by
identifying frames where local prosodic dynamics diverge most from
the temporal mean.  For each correctly classified sarcastic sample,
we compute per-frame divergence:
\begin{equation}
\begin{split}
  d_t &= \lVert\mathbf{o}_t - \bar{\mathbf{o}}\rVert_2, \\
  \bar{\mathbf{o}} &= \tfrac{1}{T}\sum\nolimits_{j=1}^{T}\mathbf{o}_j
\end{split}
\end{equation}
The predicted onset is the frame with the highest
attention-weighted divergence:
\begin{equation}
  t^{*} = \operatorname{argmax}_{t}\ (a_t \cdot d_t)
\end{equation}
where $a_t$ is the learned pooling weight from Eq.\,4 and $d_t$
the frame divergence.  This selects frames that are both
\emph{prosodically salient} (high~$a_t$) and \emph{locally
divergent} (high~$d_t$).  Since fine-grained temporal annotations
are unavailable, evaluation uses relative position within the
utterance rather than absolute timestamps.


\begin{table*}[t]
\centering
\caption{Performance of ProSarc across datasets (percentages with 95\% CI).}
\label{tab:main_results}
\renewcommand{\arraystretch}{1.3}
\footnotesize
\begin{tabular}{|l|p{2.0cm}|p{2.0cm}|p{2.0cm}|p{2.3cm}|}
\hline
\textbf{Metric} & \textbf{MUStARD++} & \textbf{MUStARD} & \textbf{PodSarc} & \textbf{MuSaG} \\
\hline
Accuracy & 73.29 [62.0–84.6] & 74.42 [69.1–79.7] & 63.60 [56.1–71.1] & 61.48 [47.6–75.4] \\
F1 Score & 75.28 [65.6–84.9] & 77.03 [72.3–81.8] & 62.89 [54.1–71.6] & 65.59 [44.5–86.7] \\
Precision & 71.62 [60.3–82.9] & 71.65 [65.1–78.2] & 64.46 [56.0–72.9] & 65.69 [48.9–82.5] \\
Recall & 79.51 [71.2–87.8] & 84.02 [74.1–93.9] & 62.40 [48.7–76.1] & 72.72 [34.9–100.0]$^{\ddagger}$ \\
MCC & 46.78 [24.3–69.3] & 49.99 [39.7–60.3] & 27.59 [12.8–42.4] & 22.12 [-7.3–51.5] \\
Cohen $\kappa$ & 46.42 [23.7–69.2] & 48.60 [37.9–59.3] & 27.20 [12.2–42.2] & 19.57 [-10.8–49.9] \\
AUC & 78.42 [69.9–86.9] & 80.58 [76.7–84.5] & 67.58 [56.8–78.4] & 63.24 [45.7–80.8] \\
\hline
\end{tabular}
\vspace{2pt}
\newline{\scriptsize $^{\ddagger}$Upper bound clipped to 100; raw Wald interval exceeds the metric range due to high fold-level variance ($N{=}213$).}
\end{table*}

\subsection{Uncertainty Estimation}
\label{ssec:uncertainty}

Predictive uncertainty is estimated via MC~dropout
\cite{uncertainity} with $T_{\text{MC}}{=}10$ stochastic forward
passes at inference.  Dropout ($p{=}0.2$) in the classification
head is retained during inference:
\begin{equation}
\label{eq:mc_dropout}
  \hat{y}_{\text{MC}} =
    \frac{1}{T_{\text{MC}}}\sum_{i=1}^{T_{\text{MC}}}\hat{y}_i,
  \quad
  \mathrm{Var} =
    \frac{1}{T_{\text{MC}}}\sum_{i=1}^{T_{\text{MC}}}
      (\hat{y}_i - \hat{y}_{\text{MC}})^{2}
\end{equation}
The mean serves as the final prediction and the variance as an
interpretability signal; the uncertainty score does not alter class
decisions.





\section{Experimental Setup}

\subsection{Datasets}


We evaluate ProSarc on four sarcasm benchmarks covering scripted, spontaneous, and cross-lingual settings (Table~\ref{tab:datasets}).  MUStARD and MUStARD++ consist of
short English clips from scripted television dialogues~\cite{ray-etal-2022-multimodal,castro-etal-2019-towards}. PodSarc contains spontaneous podcast speech with greater acoustic variability and weaker prosodic exaggeration ~\cite{li2025leveraginglargelanguagemodels}; we evaluate on a 1,000-utterance stratified random subset (fixed seed) to match the scale of other benchmarks.  MuSaG, a German-language dataset from YouTube~\cite{scott2025musagmultimodalgermansarcasm}, is included to probe whether prosodic incongruity generalises across languages despite its limited size ($N{=}213$)~\cite{crossling}.  All datasets use 5-fold cross-validation split.  All experiments use audio-only input.

\begin{table}[H]
\setlength{\tabcolsep}{3pt}
\caption{Dataset statistics (Sar./Non-Sar.; duration in seconds as [min/mean/max]).}
\label{tab:datasets}
\centering
\footnotesize
\begin{tabular}{lcccc}
\toprule
\textbf{Dataset} & \textbf{Total} & \textbf{Sar./Non-Sar.} & \textbf{Dur. (s)} & \textbf{Split} \\
\midrule
MUStARD      & 690  & 345 / 345 & [1.2/4.3/9.8]  & 5-fold CV \\
MUStARD++    & 1203 & 602 / 601 & [1.1/4.6/10.2] & 5-fold CV \\
PodSarc sample     & 1000 & 500 / 500 & [2.0/6.8/18.5] & 5-fold CV \\
MuSaG        & 213  & 119 / 94  & [1.5/5.2/14.1] & 5-fold CV \\
\bottomrule
\end{tabular}
\end{table}

\subsection{Baseline Models}

We compare against baselines that span a range of representation capacity to isolate the contribution of learned temporal features. A \emph{random baseline} (uniform class sampling) establishes the chance-level lower bound.  \emph{OpenSMILE~+~SVM}~\cite{GeMAPS} tests whether hand-crafted prosodic statistics suffice for classification.  Three SSL encoder families, namely, Wav2Vec\,2.0 (Base)~\cite{baevski2020wav2vec20frameworkselfsupervised}, HuBERT (Base, Large)~\cite{HsuBTLSM21}, and WavLM (Base, Large)~\cite{WavLM} have been used to evaluate the effect of model capacity and pretraining objective on downstream sarcasm detection. Table~\ref{tab:model_params} summarises computational cost across all configurations.



\subsection{Implementation Details}
All models are implemented in PyTorch and trained on a single NVIDIA Tesla T4 (16\,GB).  Training uses the Adam optimiser with a learning rate of $2e-5$, batch size~4, and early stopping (patience~5) on validation F1-score.  Pretrained encoders are partially fine-tuned: only the final two transformer layers are unfrozen, while all lower layers serve as frozen feature extractors~\cite{pepino21_interspeech}. We set a fixed random seed (42). Dropout ($p{=}0.2$) is applied in the classification head and retained at inference for MC~dropout sampling~\cite{uncertainity}.  We report F1-score as the primary metric, supplemented by Accuracy, Precision, Recall, MCC, Cohen's $\kappa$, and AUC-ROC to account for class imbalance. Results are reported as percentages with 95\% confidence intervals over 5-fold cross-validation.
(Table~\ref{tab:main_results}).

\section{Results and Analysis}

\subsection{Main Results}
\label{sec:main_results}

The performance of the proposed ProSarc is summarized in Table~\ref{tab:main_results}. ProSarc achieves its strong performance on the MUStARD dataset, with an accuracy of 74.4\% and an F1-score of 77.0\%, and on MUStARD++ (73.3\% and 75.3\%), which are scripted TV dialogue dataset. The performance on spontaneous datasets such as PodSarc is slightly lower (63.6\% accuracy, 62.9\% F1), which is expected due to natural conversational variability, while still demonstrating consistent MCC, Cohen's $\kappa$, and AUC values. On the multilingual MuSaG dataset, ProSarc obtains (61.5\% accuracy, 65.6\% F1), which demonstrates promising cross-lingual temporal prosodic modeling~\cite{scott2025musagmultimodalgermansarcasm}.

\begin{table}[H]
\centering
\caption{Test performance (\%) on MUStARD++.}
\label{tab:all_models}
\footnotesize
\begin{tabular}{|p{2cm}|c|c|c|c|}
\hline
Model & Accuracy & F1 Score & Precision & Recall \\
\hline
Random Baseline & 49.17 & 50.00 & 48.94 & 51.11 \\
\hline
OpenSMILE + SVM & 53.53 & 52.94 & 53.39 & 52.50 \\
\hline
Wav2Vec2-Base & 55.60 & 49.77 & 56.99 & 44.17 \\
HuBERT-Base   & 53.53 & 55.56 & 53.03 & 58.33 \\
WavLM-Base    & 56.85 & 53.15 & 57.84 & 49.17 \\
\hline
HuBERT-Large  & 68.56 & 71.59 & 67.06 & 77.68 \\
WavLM-Large   & \textbf{73.29} & \textbf{75.28} & \textbf{71.62} & \textbf{79.51} \\
\hline
\end{tabular}
\end{table}

Hypothesis testing confirms that enabling incongruity-aware training leads to consistent improvements in sarcasm detection. For 10 different runs with different random seeds, the experimental WavLM-Large model performs better than the baseline in all runs, and all differences are positive ($\Delta$F1 = 0.0311 on average). A Wilcoxon signed-rank test yields $p = 0.00195$, and a paired $t$-test yields $p = 0.0031$, both of which strongly support $H_1$. The result is extremely large, with a Cohen’s $d$ statistic of 1.51. McNemar’s test on paired test-set predictions yields $p = 0.779$, indicating no significant difference in instance-level error patterns despite consistent aggregate gains. In addition, a 95\% bootstrap confidence interval for the mean F1 difference is $[0.0180,\, 0.0442]$, which further supports the importance of temporal prosodic incongruity in improving the detection of sarcasm.

As seen in Table~\ref{tab:all_models} from a single evaluation run, larger self-supervised encoders consistently outperform smaller models and traditional baselines on MUStARD++, with WavLM-Large achieving the best overall F1-score and recall. Table~\ref{tab:comparison} illustrates that ProSarc outperforms the previous best audio-only approaches on all scripted, spontaneous, and cross-lingual benchmarks~\cite{crossling}. To ensure a fair comparison, we report audio-only results from prior work: for B\v{a}roiu et al. (2023)\cite{electronics12030666} and Gao et al. (2025)\cite{mused}, the reported F1-scores correspond to the audio modality alone (averaging “with context” and “without context” where applicable), whereas for Ray et al. (2022)\cite{ray-etal-2022-multimodal} and Tiwari et al. (2023)\cite{tiwari2023}, we use the best reported audio-only performance.

\begin{table}[H]
\centering
\caption{Comparison with prior audio-only sarcasm detection work in terms of F1-score (\%).}
\label{tab:comparison}
\scriptsize
\renewcommand{\arraystretch}{1.15}
\begin{tabular}{|l|p{1.4cm}|>{\centering\arraybackslash}p{0.7cm}|c|}
\hline
\textbf{Prior Work} & \textbf{Dataset} & \textbf{Prior Result} & \textbf{ProSarc} \\
\hline
B\v{a}roiu et al. (2023)~\cite{electronics12030666} & MUStARD & 60.1 & \textbf{77.03} \\
Gao et al. (2025) ~\cite{mused} & MUStARD & 67.9 & \textbf{77.03} \\
\hline
Ray et al. (2022)~\cite{ray-etal-2022-multimodal} & MUStARD++ & 64.5 & \textbf{75.3} \\

Tiwari et al., 2023~\cite{tiwari2023} & MUStARD++ & 66.6 & \textbf{75.3} \\
\hline
\end{tabular}
\end{table}

\subsection{Exploratory Temporal Onset Analysis}

\begin{figure}[htbp]
    \centering
    \includegraphics[width=\columnwidth,height=\columnwidth,keepaspectratio]{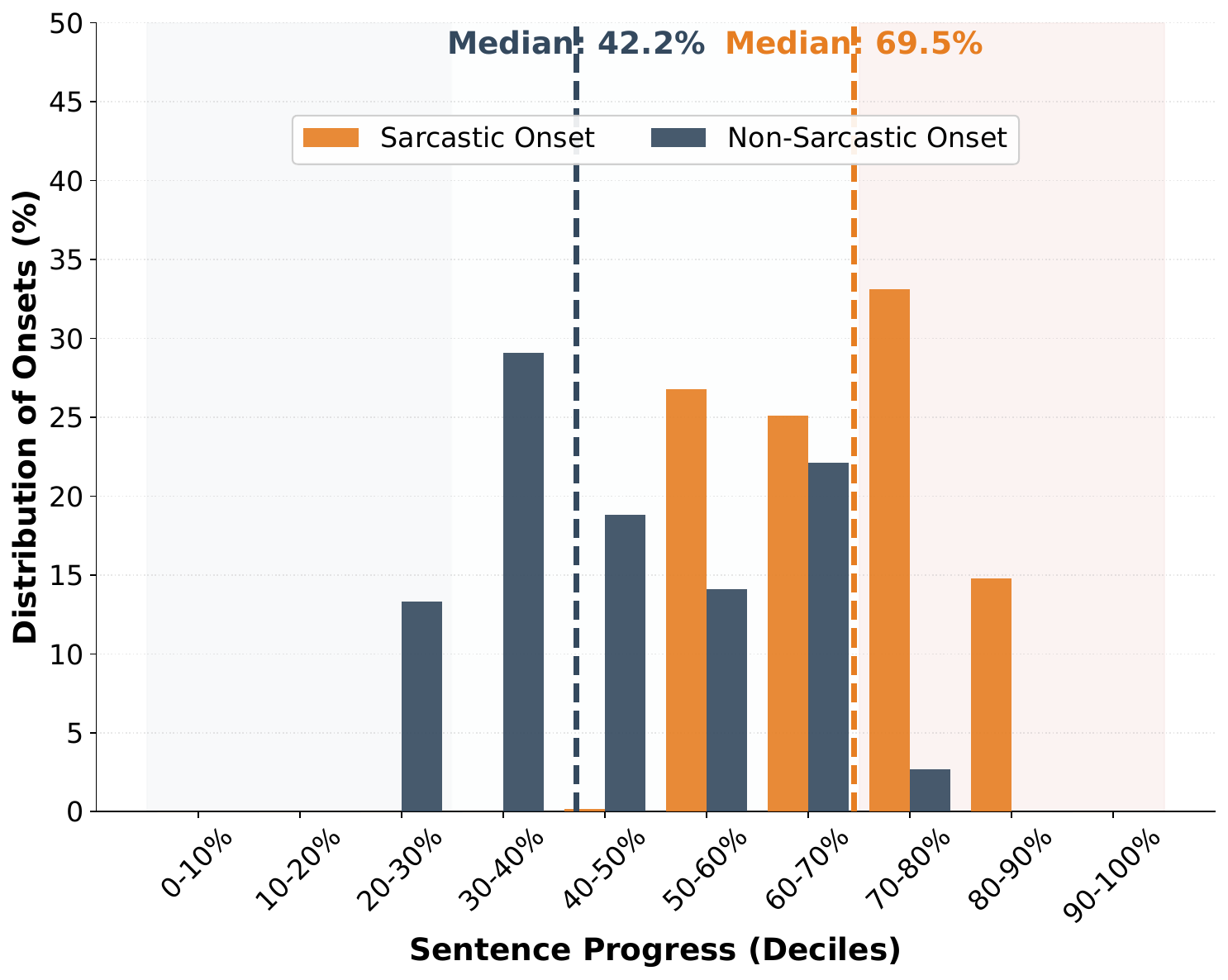}
    \caption{Decile-wise distribution of sarcasm onset}
    \label{fig:sarcasm_onset}
\end{figure}

As an exploratory experiment, we now investigate the distribution of model-predicted sarcasm onsets in utterances. The predicted onsets of sarcastic utterances show a clear late-utterance clustering, with mean relative position 68.2\%, median 69.5\%, and standard deviation 10.0\% of utterance duration (Figure~\ref{fig:sarcasm_onset}). The most frequent onsets lie in the 70-80\% position (33.1\%).
Non-sarcastic utterances, by contrast, tend to appear earlier in the utterance, with mean relative position 46.2\%, median 42.2\%, and standard deviation 14.1\%. Their most frequent onsets cluster in the 30--40\% region (29.1\%). Importantly, the onset distribution for non-sarcastic utterances does not exhibit a comparable late-utterance peak, but instead shows a broader and earlier concentration, suggesting that the 70–80\% clustering observed for sarcastic utterances is not merely a generic attention bias toward utterance-final positions. Note that this distribution is to be understood as the learned attention pattern of the model, not as a validated temporal perception of sarcasm, and is included to point out possible temporal patterns that could be explored in future annotation studies.

\subsection{Uncertainty Analysis}

Predictive uncertainty is quantified as the mean variance of the model's output probabilities across the test set, computed by retaining Monte Carlo (MC) dropout ($p{=}0.2$) in the classification head during inference and aggregating $T_{\text{MC}}{=}10$ stochastic forward passes per sample (Eq.~\ref{eq:mc_dropout}).

Table~\ref{tab:uncertainty} reports the mean test-set uncertainty for each encoder configuration. A consistent trend emerges: larger self-supervised encoders yield \emph{lower} predictive variance. HuBERT-Large achieves the lowest uncertainty ($3.7 \times 10^{-3}$), followed by WavLM-Large ($4.4 \times 10^{-3}$), while Base-scale encoders exhibit roughly $1.2$--$1.8\times$ higher variance ($5.4$--$6.8 \times 10^{-3}$). This pattern indicates that higher-capacity encoders produce representations that are more \emph{robust} to stochastic dropout perturbations, consistent with prior findings that larger pretrained models learn smoother, better-generalising feature spaces~\cite{uncertainity, lakshminarayanan2017simplescalablepredictiveuncertainty}.

\begin{table}[H]
\centering
\caption{Predictive uncertainty (mean MC dropout variance, $T_{\text{MC}}{=}10$) across audio encoders on MUStARD++. Lower values indicate greater robustness to dropout perturbation.}
\label{tab:uncertainty}
\renewcommand{\arraystretch}{1.2}
\footnotesize
\begin{tabular}{lcc}
\toprule
\textbf{Model} & \textbf{Mean Variance ($\downarrow$)} & \textbf{F1 (\%)} \\
\midrule
OpenSMILE + SVM & \textemdash & 52.94 \\
Wav2Vec2-Base & $6.8 \times 10^{-3}$ & 49.77 \\
HuBERT-Base & $5.5 \times 10^{-3}$ & 55.56 \\
WavLM-Base & $5.4 \times 10^{-3}$ & 53.15 \\
HuBERT-Large & $3.7 \times 10^{-3}$ & 71.59 \\
WavLM-Large & $4.4 \times 10^{-3}$ & 75.28 \\
\bottomrule
\end{tabular}
\end{table}

Notably, WavLM-Large achieves the highest classification performance (F1\,=\,75.28, Table~\ref{tab:main_results}) yet exhibits slightly higher uncertainty than HuBERT-Large ($4.4 \times 10^{-3}$ vs.\ $3.7 \times 10^{-3}$). This suggests that the two metrics capture complementary properties: F1 reflects aggregate discriminative accuracy, whereas MC dropout variance reflects local sensitivity of the posterior around the decision boundary~\cite{kendall}. A model may produce well-calibrated uncertainty estimates while achieving marginally different peak performance, reinforcing the value of reporting uncertainty alongside standard classification metrics, particularly for subjective tasks such as sarcasm detection where ground-truth labels themselves carry annotator disagreement~\cite{Rockwell2000}. From a deployment perspective, high MC dropout variance could serve as a natural trigger to route samples to a multimodal pipeline or flag them for human review, as explored in Section~\ref{sec:human_eval}. The OpenSMILE+SVM baseline produces hard classification boundaries without probabilistic outputs compatible with MC dropout and is therefore excluded.

\subsection{Human Evaluation}
\label{sec:human_eval}

Sarcasm perception integrates prosodic, lexical, and visual cues
simultaneously~\cite{Rockwell2000,cheang2009,Bryant11102010}.  We
employ three independent annotators under two modality conditions:
R1 and R2 label clips from \emph{audio only}, providing
modality-matched inter-annotator agreement, while R3 receives the
\emph{full video} and supplies temporal onset annotations
(Section~\ref{ssec:onset}).  All three were experienced in speech and
affective analysis; none received task-specific training.  The
evaluation targets the 50~samples with highest predictive uncertainty
(MC~dropout variance $\geq 6.77\times10^{-3}$; confidence within
0.07 of the decision boundary).

\noindent\textbf{Protocol.}
R1 and R2 independently assigned binary labels
(\textsc{sarc}/\textsc{no\,sarc}) from audio alone, with clip order
randomised per annotator and no access to video, model predictions,
or each other's labels.  R3 viewed the full video and marked the
start, peak, and end of the perceived sarcastic segment as percentages of
utterance duration, based on visible facial cues, delivery emphasis, and
gestural timing, with free-text cue descriptions.  R3 identified
at least one sarcastic interval in all~50 clips.

\noindent\textbf{Audio-only agreement.}
R1 labelled 33 clips as sarcastic; R2 labelled 27 (Cohen's
$\kappa{=}0.34$, Krippendorff's $\alpha{=}0.34$, raw agreement
68.0\%; Table~\ref{tab:human_eval}).  Three-way $\alpha$ across R1,
R2, and the model is 0.33, dropping to 0.16 when R3's multimodal
labels are included---confirming that the audio-only raters and model
occupy a coherent annotation space distinct from multimodal
perception.  The moderate $\kappa$ is consistent with known
subjectivity of audio-only sarcasm judgement~\cite{Rockwell2000} and
expected given that these are the model's most uncertain predictions.

\begin{table}[H]
\centering
\caption{Human evaluation on the 50 most uncertain predictions.
R1, R2: audio-only binary.  R3: multimodal temporal annotation.}
\label{tab:human_eval}
\footnotesize
\renewcommand{\arraystretch}{1.15}
\begin{tabular}{|l|c|c|c|}
\hline
\textbf{Metric}
  & \textbf{R1 (audio)}
  & \textbf{R2 (audio)}
  & \textbf{R3 (mm)} \\
\hline
Labels (S\,/\,NS)
  & 33\,/\,17
  & 27\,/\,23
  & 50\,/\,0$^{\dagger}$ \\
Agree w/ model (\%)
  & 74.0
  & 70.0
  & 84.0 \\
$\kappa$ vs.\ model
  & 0.34
  & 0.37
  & --- \\
\hline
\multicolumn{2}{|l|}{$\kappa$ (R1 vs R2)}
  & \multicolumn{2}{c|}{0.34} \\
\multicolumn{2}{|l|}{$\alpha$ (R1, R2, model)}
  & \multicolumn{2}{c|}{0.33} \\
\multicolumn{2}{|l|}{$\alpha$ (all four)}
  & \multicolumn{2}{c|}{0.16} \\
\hline
\end{tabular}
\vspace{1pt}
\newline{\scriptsize $^{\dagger}$Sarcastic cues identified in all 50
clips under multimodal access.}
\end{table}

\begin{figure*}[t]
  \centering
  \includegraphics[width=0.95\linewidth,height=0.25\textheight]{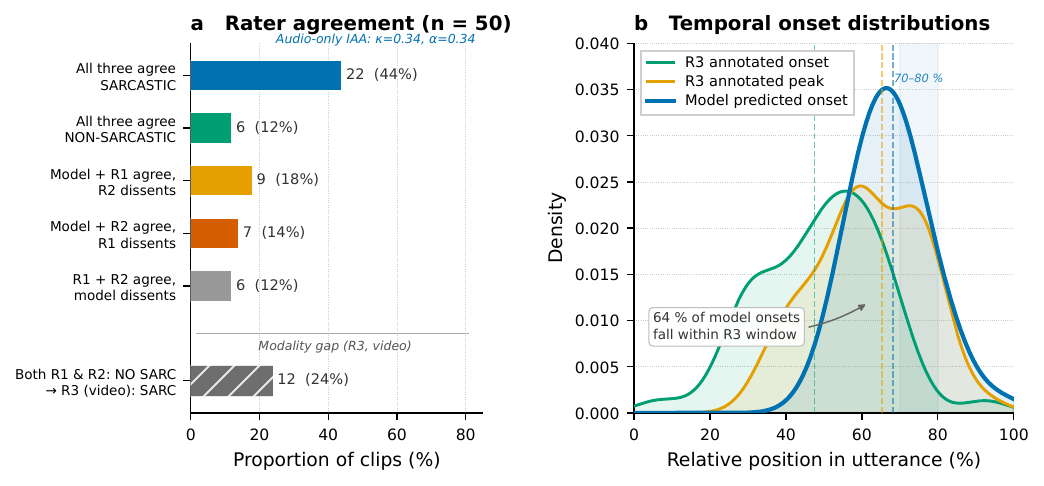}
  \caption{Human evaluation on the 50 most uncertain predictions.
    \textbf{(a)}~Audio-condition agreement: R1, R2 (audio-only), and
    model; hatched bar shows the modality gap (both audio raters
    label non-sarcastic, R3 with video identifies sarcasm).
    \textbf{(b)}~Temporal onset KDEs: R3-annotated onset and peak
    versus model prediction.  Shaded region: modal decile
    (70--80\%).}
  \label{fig:human_eval}
\end{figure*}

\noindent\textbf{Model--human agreement.}
Model--R1 agreement is 74.0\% ($\kappa{=}0.34$), model--R2 is 70.0\%
($\kappa{=}0.37$), and model--R3 (multimodal) is 84.0\%.  The
progressive increase traces the \emph{perceptual cost of modality
restriction}.  Figure~\ref{fig:human_eval}a decomposes disagreement:
all three audio-condition raters converge in 56\% of clips (44\%
sarcastic, 12\% non-sarcastic); no clip produced three-way
disagreement.  Notably, 12~clips (24\%) were labelled non-sarcastic
by \emph{both} audio raters yet sarcastic by R3, directly quantifying
the perceptual territory recoverable only through visual channels.

\noindent\textbf{Temporal onset validation.}
We use R3's multimodal markings as the reference standard, since
temporal sarcasm onset integrates cues across modalities.  The
50~clips are drawn from MUStARD/MUStARD++, consisting of scripted
dialogues from \emph{The Big Bang Theory} and similar sitcoms; R3's
temporal markings therefore reflect character-specific delivery styles
(e.g.\ Sheldon's deadpan monotone, Penny's exaggerated inflection)
and visual cues such as facial mugging and gestural timing that
jointly shape the perceived onset of sarcastic intent.
Figure~\ref{fig:human_eval}b overlays R3-annotated onset (mean\,=\,46.9\%), R3-annotated
peak (mean\,=\,64.9\%), and model-predicted onset (mean\,=\,68.2\%).
The model's predicted position aligns more closely with R3's
\emph{peak} than with R3's onset, suggesting that the
$\operatorname{argmax}_t(a_t \cdot d_t)$ mechanism identifies the
moment of maximum prosodic divergence rather than the first
perceptible sarcastic cue.  The model's modal decile (70--80\%) falls
just rightward of R3's peak decile (60--70\%), and its narrower
spread (std\,=\,10.0\% vs.\ 17.6\% for R3 peak and 20.1\% for R3
onset) indicates that it captures a central tendency of the sarcasm
peak while underrepresenting the full range of onset positions.
Moreover, 64\% of model-predicted onsets fall within R3's annotated
window $[\text{Start}_\%,\,\text{End}_\%]$---non-trivial given the
absence of temporal supervision during training---suggesting that the
learned attention weights track a prosodic transition that human
listeners independently perceive as the locus of sarcastic intent.

\noindent\textbf{Qualitative observations.}
R3's free-text descriptions (43/50 clips) identify facial expressions
(53\%), vocal delivery style (40\%), and pointed intent markers (44\%)
as the dominant cue categories (non-exclusive; individual clips often
combine multiple cue types).  The 12~clips labelled non-sarcastic by both R1
and R2 were described by R3 with character-grounded notes such as
``Sheldon's pursed-lip matter-of-fact jab'' and ``Airline agent's
polite-smile line landing as sarcasm''---cues tied to recognisable
character personas and performance styles accessible through vision
but largely absent from audio.

\noindent\textbf{Scope.}
The design provides modality-matched IAA ($\kappa{=}0.34$) and an
independent multimodal reference, but rests on three annotators and
50~uncertain predictions.  Replication with larger pools, confident
predictions, and standardised temporal guidelines remains future work.

\subsection{Ablation Study}

The ablation study on MUStARD++ is performed to measure the individual contribution of each major architectural component. For each ablated variant, the model is trained and tested using the same training setup and a fixed random seed 42 to facilitate a fair comparison. Although the results are reported based on a single deterministic run, this study aims to decouple the effect of individual components rather than measuring the performance variability.

\begin{table}[H]
\centering
\caption{Ablation study on MUStARD++ showing the impact of each component on F1-score.}
\label{tab:ablation}
\scriptsize
\renewcommand{\arraystretch}{1.15}
\begin{tabular}{lccc}
\toprule
\textbf{Model Variant} & \textbf{F1 (\%)} & $\boldsymbol{\Delta}$\textbf{F1 (abs.)} & $\boldsymbol{\Delta}$\textbf{F1 (rel.)} \\
\midrule
Full ProSarc & 70.79 & --- & --- \\
w/o Incongruity Analyzer & 67.69 & $-3.10$ & $-4.4\%$ \\
w/o Temporal Prosody Encoder & 69.31 & $-1.48$ & $-2.1\%$ \\
w/o Global Emotional Encoder & 68.40 & $-2.39$ & $-3.4\%$ \\
\bottomrule
\end{tabular}
\end{table}

Based on the results presented in Table~\ref{tab:ablation}, the removal of the Prosodic Incongruity Analyzer has the most significant impact on the model’s performance, causing a F1-score drop from 70.79 to 67.69 ($\Delta$F1 = $-3.10$, $-4.4\%$ relative), which clearly indicates that the modeling of prosodic incongruity is a crucial aspect of the model. The removal of the Temporal Prosody Encoder also has a steady negative impact ($\Delta$F1 = $-1.48$, $-2.1\%$), which indicates that the temporal aggregation of prosodic information is another complementary aspect of the model. Similarly, the removal of the Global Emotional Encoder also has a significant negative impact ($\Delta$F1 = $-2.39$, $-3.4\%$), which clearly indicates the importance of the emotional context of the utterance.

\begin{figure}[H]
    \centering
    \includegraphics[width=0.8\columnwidth,height=0.25\textheight]{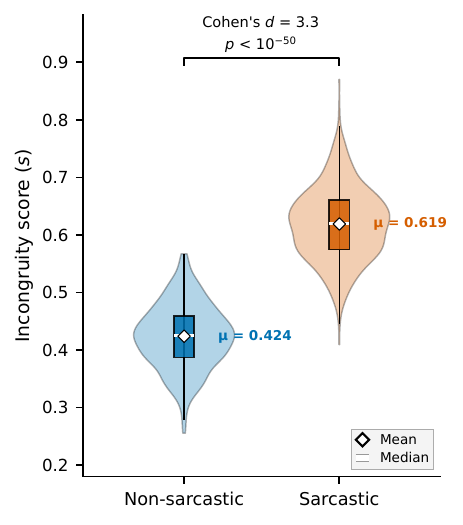}
    \caption{Distribution of learned incongruity score $s$
    for sarcastic vs.\ non-sarcastic utterances on MUStARD++
    (5-fold CV). White diamond: mean; white line: median.}
    \label{fig:incongruity_dist}
\end{figure}

In summary, the ablation experiments show that all components make a significant contribution to performance, and the Prosodic Incongruity Analyzer has the greatest impact. This is consistent with the statistical analysis of multiple runs discussed in the previous section, and it again emphasizes the importance of modeling prosodic incongruity in the task of sarcasm detection.

Figure~\ref{fig:incongruity_dist} confirms that the learned
incongruity score captures a meaningful class-level signal.
Across 5-fold cross-validation, sarcastic utterances produce
systematically higher $s$ values ($0.619 \pm 0.063$) than
non-sarcastic ones ($0.424 \pm 0.056$), with a large effect
size (Cohen's $d \approx 3.3$, $p < 10^{-50}$).

\subsection{Discussion}
\label{sec:discussion}

\noindent\textbf{What does the incongruity score capture?}
The ablation study identifies the Prosodic Incongruity Analyzer as the single most impactful component ($\Delta$F1$=-3.10$, $-4.4\%$ relative), and the 10-run statistical test confirms this with a large effect size (Cohen's $d{=}1.51$). Because $s$ is learned end-to-end without auxiliary supervision, its behaviour warrants scrutiny. The fusion mechanism in Eq.~6 uses $s$ to interpolate between local and global representations, while Eq.~7 concatenates the fused output with the original $\mathbf{z}$, creating a skip connection that preserves both pathways. This design ensures the classifier has access to complementary information, but also means the model is not forced to route information exclusively through the gating mechanism. Despite this, ablating the incongruity analyzer produces the largest single-component performance drop across all variants, suggesting that the learned $s$ provides a discriminative signal beyond what the skip connection alone delivers. The distribution of $s$ values (Figure~\ref{fig:incongruity_dist}) confirms this: sarcastic utterances produce systematically higher incongruity scores ($0.619 \pm 0.063$ vs.\ $0.424 \pm 0.056$; Cohen's $d \approx 3.3$), validating that $s$ functions as a meaningful incongruity detector rather than collapsing to a fixed gating value.

\noindent\textbf{Temporal dynamics: model onset as peak detector.}
Cross-referencing model predictions with R3's frame-level multimodal annotations reveals a consistent temporal relationship. The mean human-annotated sarcasm onset (46.9\%, std=20.1\%) precedes the model's predicted onset (68.2\%, std=10.0\%), whereas the human-annotated sarcasm peak (64.9\%, std=17.6\%) closely matches it. This suggests that the $\mathrm{argmax}_t(a_t \cdot d_t)$ mechanism in Eq.~10 captures something closer to the \emph{peak} of sarcastic expression rather than the perceptual onset. This interpretation is consistent with the psycholinguistic literature: Rockwell~\cite{Rockwell2000} and Bryant~\cite{Bryant11102010} describe prosodic exaggeration and contrastive stress as key sarcasm markers, which would reach maximum divergence from the emotional baseline at the expressive peak rather than at the initial delivery shift.

The model's narrower standard deviation (10.0\% vs.\ 20.1\% for onset and 17.6\% for peak) indicates that it captures a central tendency in prosodic timing but underestimates the full range of sarcastic delivery patterns. R3's annotations span from 8\% (early pointed jab) to 92\% (late-utterance smirk), reflecting genuine diversity in sarcasm realisation that a single-onset prediction cannot fully represent. Two clips exhibiting secondary sarcastic beats (1\_1560, 1\_1973) further suggest that sarcasm can manifest as multiple prosodic events within a single utterance, a phenomenon not captured by the current single-onset architecture.

\noindent\textbf{The modality ceiling for audio-only detection.}
The 12 clips (24\%) labelled non-sarcastic by both audio-only annotators but identified as sarcastic by R3 under multimodal access establish an empirical lower bound on the irreducible error of any audio-only system for this sarcasm subtype. R3's free-text descriptions for these clips reveal predominantly visual cues: pursed-lip looks, side-glances, and polite-smile deliveries that are character-specific and accessible through vision but largely absent from the audio signal. These clips are also significantly shorter than the remaining samples (mean duration 4.8\,s vs.\ 6.9\,s, $t{=}{-}2.07$, $p{=}0.043$), providing less temporal signal for the prosody encoder, and have narrower sarcasm spans (28.5\% vs.\ 33.9\%).

This modality gap reframes the interpretation of aggregate results. The PodSarc results (F1\,=\,62.9\%) should be interpreted in this light: spontaneous conversational speech likely contains a higher proportion of visually-signalled sarcasm, placing a lower ceiling on audio-only detection than scripted television dialogue.

\noindent\textbf{Uncertainty aligns with perceptual ambiguity.}
The 50 clips selected for human evaluation have confidence scores ranging from 0.500 to 0.571 (mean distance from the decision boundary: 0.028), confirming that MC dropout uncertainty correctly identifies a zone of genuine perceptual ambiguity where inter-annotator agreement is moderate ($\kappa{=}0.34$) and 32\% of clips produce rater disagreement.

Within this narrow band, a monotonic confidence gradient tracks human consensus among the model's sarcastic predictions: clips where both raters agree \textsc{sarc} have the highest mean confidence (0.533), those with split rater judgement have intermediate confidence (0.529), and clips where both raters disagree with the model have the lowest (0.527). Although the absolute range is narrow, this gradient suggests that the model's posterior probability encodes meaningful information about perceptual consensus. Notably, in all 16 cases of inter-rater disagreement, the model agrees with at least one human annotator, never producing a prediction that both raters simultaneously reject. From a deployment perspective, high MC dropout variance could trigger fallback to a multimodal pipeline or flag items for human review~\cite{uncertainity,kendall}.

\noindent\textbf{Dataset-specific error patterns.}
Performance degradation across datasets traces to distinct failure modes. On PodSarc, the lower F1 (62.9\%) relative to scripted benchmarks reflects acoustic noise, speaker variability, and subtler sarcastic delivery. On MUStARD and MUStARD++, errors concentrate on context-dependent sarcasm where intent relies on dialogue history or situational irony that an utterance-level model cannot access. The six clips where the model predicts \textsc{sarcastic} but both audio-only raters disagree have mean confidence (0.527) indistinguishable from the overall sample mean (0.528), indicating these are genuinely ambiguous instances rather than confident misclassifications.

On MuSaG ($N{=}213$), the wide confidence intervals (F1: 44.5--86.7\%, MCC CI crossing zero) reflect both the small dataset size and the English-centric pretraining of the SSL encoders. These results should be interpreted as preliminary cross-lingual indications rather than evidence of robustness. Future work should leverage larger multilingual datasets and domain-adaptive pretraining to evaluate cross-lingual generalisation more rigorously.

\subsection{Limitations}

ProSarc depends solely on audio and does not exploit semantic cues, which may limit detection when sarcasm is primarily conveyed through words, as leveraged by multimodal approaches~\cite{Gao2025SpokenIJ}. Temporal sarcasm onset is estimated via weak supervision and shows distributional alignment with human temporal annotations (Section~\ref{sec:human_eval}), though per-sample validation remains limited to 50~clips and the temporal annotations were produced under multimodal (audio + video) conditions rather than audio-only. Furthermore, the model processes utterances in isolation, ignoring conversational context that often influences multi-turn sarcasm.

Although ProSarc generates confidence estimates for comparative and diagnostic purposes, the confidence values reflect relative predictive uncertainty rather than calibrated probabilities; they are useful for model analysis and flagging uncertain cases but are not optimized for probabilistic calibration. Additionally, the single-onset prediction ($t^{*}$ in Eq.~10) cannot represent clips containing multiple sarcastic beats, as observed in two of the 50 evaluated clips.

More broadly, human evaluation reveals that sarcasm perception is modality-dependent: annotators with access to visual cues identify sarcasm in cases that audio-only assessment misses (Krippendorff's $\alpha{=}0.16$ across all four raters including the multimodal annotator, compared to $\alpha{=}0.33$ among the three audio-condition raters alone). This gap suggests that audio-only sarcasm detection operates on a subset of perceptible sarcastic cues, and that modality restriction, not just task subjectivity, places a ceiling on achievable agreement.

\section{Conclusion}

We presented ProSarc, an audio-only framework that models sarcasm
as temporal prosodic incongruity between local dynamics and a global
emotional baseline.  On MUStARD++ (F1\,=\,75.3), ProSarc
outperforms prior audio-only methods, and generalises to spontaneous
speech (PodSarc, F1\,=\,62.9) and cross-lingual data (MuSaG,
F1\,=\,65.6), with statistical confirmation across ten runs
(Wilcoxon $p{=}0.002$, Cohen's $d{=}1.51$).  The learned
incongruity score separates sarcastic from non-sarcastic utterances
with a large effect size ($d \approx 3.3$), validating that the
model captures a meaningful prosodic signal.

Human evaluation on the 50 most uncertain predictions reveals that
model-predicted onsets align with human-annotated sarcasm peaks
(64\% within annotated windows), and that 24\% of clips require
visual cues inaccessible to any audio-only system---establishing
an empirical modality ceiling.  MC~dropout uncertainty tracks
inter-annotator disagreement, suggesting a natural trigger for
multimodal fallback in deployment.

Future work should extend temporal modelling to multi-event sarcasm,
incorporate visual and contextual signals to address the modality
ceiling, and evaluate on larger cross-lingual corpora with
domain-adaptive pretraining.

\section{Generative AI Use Disclosure}
The authors used LLM-based tool (ChatGPT) solely for editorial assistance: improving clarity, grammar, conciseness, and stylistic consistency of the manuscript text. No generative AI was used to produce the research ideas, experimental design, code, analyses, results, or scientific claims, which are entirely the authors' own. All AI-assisted edits were reviewed and verified by the authors, who take full responsibility for the content of the paper.

\bibliographystyle{IEEEtran}
\bibliography{main}

\begin{thebibliography}{10}
\providecommand{\url}[1]{#1}
\csname url@samestyle\endcsname
\providecommand{\newblock}{\relax}
\providecommand{\bibinfo}[2]{#2}
\providecommand{\BIBentrySTDinterwordspacing}{\spaceskip=0pt\relax}
\providecommand{\BIBentryALTinterwordstretchfactor}{4}
\providecommand{\BIBentryALTinterwordspacing}{\spaceskip=\fontdimen2\font plus
\BIBentryALTinterwordstretchfactor\fontdimen3\font minus
  \fontdimen4\font\relax}
\providecommand{\BIBforeignlanguage}[2]{{%
\expandafter\ifx\csname l@#1\endcsname\relax
\typeout{** WARNING: IEEEtran.bst: No hyphenation pattern has been}%
\typeout{** loaded for the language `#1'. Using the pattern for}%
\typeout{** the default language instead.}%
\else
\language=\csname l@#1\endcsname
\fi
#2}}
\providecommand{\BIBdecl}{\relax}
\BIBdecl

\bibitem{Rockwell2000}
\BIBentryALTinterwordspacing
P.~Rockwell, ``Lower, slower, louder: Vocal cues of sarcasm,'' \emph{Journal of
  Psycholinguistic Research}, vol.~29, no.~5, pp. 483--495, 2000. [Online].
  Available: \url{https://doi.org/10.1023/A:1005120109296}
\BIBentrySTDinterwordspacing

\bibitem{cheang2009}
\BIBentryALTinterwordspacing
H.~S. Cheang and M.~D. Pell, ``Acoustic markers of sarcasm in cantonese and
  english,'' \emph{The Journal of the Acoustical Society of America}, vol. 126,
  no.~3, pp. 1394--1405, 09 2009. [Online]. Available:
  \url{https://doi.org/10.1121/1.3177275}
\BIBentrySTDinterwordspacing

\bibitem{Bryant11102010}
\BIBentryALTinterwordspacing
G.~A. Bryant, ``Prosodic contrasts in ironic speech,'' \emph{Discourse
  Processes}, vol.~47, no.~7, pp. 545--566, 2010. [Online]. Available:
  \url{https://doi.org/10.1080/01638530903531972}
\BIBentrySTDinterwordspacing

\bibitem{ray-etal-2022-multimodal}
\BIBentryALTinterwordspacing
A.~Ray, S.~Mishra, A.~Nunna, and P.~Bhattacharyya, ``A multimodal corpus for
  emotion recognition in sarcasm,'' in \emph{Proceedings of the Thirteenth
  Language Resources and Evaluation Conference}.\hskip 1em plus 0.5em minus
  0.4em\relax Marseille, France: European Language Resources Association, Jun.
  2022, pp. 6992--7003. [Online]. Available:
  \url{https://aclanthology.org/2022.lrec-1.756/}
\BIBentrySTDinterwordspacing

\bibitem{castro-etal-2019-towards}
\BIBentryALTinterwordspacing
S.~Castro, D.~Hazarika, V.~P{\'e}rez-Rosas, R.~Zimmermann, R.~Mihalcea, and
  S.~Poria, ``Towards multimodal sarcasm detection (an {\_}{O}bviously{\_}
  perfect paper),'' in \emph{Proceedings of the 57th Annual Meeting of the
  Association for Computational Linguistics}, A.~Korhonen, D.~Traum, and
  L.~M{\`a}rquez, Eds.\hskip 1em plus 0.5em minus 0.4em\relax Florence, Italy:
  Association for Computational Linguistics, Jul. 2019, pp. 4619--4629.
  [Online]. Available: \url{https://aclanthology.org/P19-1455/}
\BIBentrySTDinterwordspacing

\bibitem{poria-etal-2019-meld}
\BIBentryALTinterwordspacing
S.~Poria, D.~Hazarika, N.~Majumder, G.~Naik, E.~Cambria, and R.~Mihalcea,
  ``{MELD}: A multimodal multi-party dataset for emotion recognition in
  conversations,'' in \emph{Proceedings of the 57th Annual Meeting of the
  Association for Computational Linguistics}, A.~Korhonen, D.~Traum, and
  L.~M{\`a}rquez, Eds.\hskip 1em plus 0.5em minus 0.4em\relax Florence, Italy:
  Association for Computational Linguistics, Jul. 2019, pp. 527--536. [Online].
  Available: \url{https://aclanthology.org/P19-1050/}
\BIBentrySTDinterwordspacing

\bibitem{schuller2018}
\BIBentryALTinterwordspacing
B.~W. Schuller, ``Speech emotion recognition: two decades in a nutshell,
  benchmarks, and ongoing trends,'' \emph{Commun. ACM}, vol.~61, no.~5, p.
  90–99, Apr. 2018. [Online]. Available:
  \url{https://doi.org/10.1145/3129340}
\BIBentrySTDinterwordspacing

\bibitem{gao22}
X.~Gao, S.~Nayak, and M.~Coler, ``Deep cnn-based inductive transfer learning
  for sarcasm detection in speech,'' in \emph{Proc. Interspeech 2022}, 09 2022,
  pp. 2323--2327.

\bibitem{vaswani2023attentionneed}
\BIBentryALTinterwordspacing
A.~Vaswani, N.~Shazeer, N.~Parmar, J.~Uszkoreit, L.~Jones, A.~N. Gomez, L.~u.
  Kaiser, and I.~Polosukhin, ``Attention is all you need,'' in \emph{Advances
  in Neural Information Processing Systems}, I.~Guyon, U.~V. Luxburg,
  S.~Bengio, H.~Wallach, R.~Fergus, S.~Vishwanathan, and R.~Garnett, Eds.,
  vol.~30.\hskip 1em plus 0.5em minus 0.4em\relax Curran Associates, Inc.,
  2017. [Online]. Available:
  \url{https://proceedings.neurips.cc/paper_files/paper/2017/file/3f5ee243547dee91fbd053c1c4a845aa-Paper.pdf}
\BIBentrySTDinterwordspacing

\bibitem{uncertainity}
Y.~Gal and Z.~Ghahramani, ``Dropout as a bayesian approximation: Representing
  model uncertainty in deep learning,'' \emph{Proceedings of The 33rd
  International Conference on Machine Learning}, 06 2015.

\bibitem{funfgeld25_interspeech}
S.~Fünfgeld, A.~Braun, and K.~Zahner-Ritter, ``{Are You Being Sarcastic?
  Prosodic Cues to Irony Perception in German},'' in \emph{{Interspeech 2025}},
  2025, pp. 5373--5377.

\bibitem{li2024functionaltradeoffprosodicsemantic}
Z.~Li, X.~Gao, Y.~Zhang, S.~Nayak, and M.~Coler, ``A functional trade-off
  between prosodic and semantic cues in conveying sarcasm,'' in \emph{Proc.
  Interspeech 2024}, 2024, pp. 1070--1074.

\bibitem{joshi17}
\BIBentryALTinterwordspacing
A.~Joshi, P.~Bhattacharyya, and M.~J. Carman, ``Automatic sarcasm detection: A
  survey,'' \emph{ACM Comput. Surv.}, vol.~50, no.~5, Sep. 2017. [Online].
  Available: \url{https://doi.org/10.1145/3124420}
\BIBentrySTDinterwordspacing

\bibitem{ghosh-veale-2016-fracking}
\BIBentryALTinterwordspacing
A.~Ghosh and T.~Veale, ``Fracking sarcasm using neural network,'' in
  \emph{Proceedings of the 7th Workshop on Computational Approaches to
  Subjectivity, Sentiment and Social Media Analysis}, A.~Balahur, E.~van~der
  Goot, P.~Vossen, and A.~Montoyo, Eds.\hskip 1em plus 0.5em minus 0.4em\relax
  San Diego, California: Association for Computational Linguistics, Jun. 2016,
  pp. 161--169. [Online]. Available: \url{https://aclanthology.org/W16-0425/}
\BIBentrySTDinterwordspacing

\bibitem{khodak-etal-2018-large}
\BIBentryALTinterwordspacing
M.~Khodak, N.~Saunshi, and K.~Vodrahalli, ``A large self-annotated corpus for
  sarcasm,'' in \emph{Proceedings of the Eleventh International Conference on
  Language Resources and Evaluation ({LREC} 2018)}, N.~Calzolari, K.~Choukri,
  C.~Cieri, T.~Declerck, S.~Goggi, K.~Hasida, H.~Isahara, B.~Maegaard,
  J.~Mariani, H.~Mazo, A.~Moreno, J.~Odijk, S.~Piperidis, and T.~Tokunaga,
  Eds.\hskip 1em plus 0.5em minus 0.4em\relax Miyazaki, Japan: European
  Language Resources Association (ELRA), May 2018. [Online]. Available:
  \url{https://aclanthology.org/L18-1102/}
\BIBentrySTDinterwordspacing

\bibitem{TsaiBLKMS19}
\BIBentryALTinterwordspacing
Y.~H. Tsai, S.~Bai, P.~P. Liang, J.~Z. Kolter, L.~Morency, and
  R.~Salakhutdinov, ``Multimodal transformer for unaligned multimodal language
  sequences,'' in \emph{Proceedings of the 57th Conference of the Association
  for Computational Linguistics, {ACL} 2019, Florence, Italy, July 28- August
  2, 2019, Volume 1: Long Papers}, A.~Korhonen, D.~R. Traum, and
  L.~M{\`{a}}rquez, Eds.\hskip 1em plus 0.5em minus 0.4em\relax Association for
  Computational Linguistics, 2019, pp. 6558--6569. [Online]. Available:
  \url{https://doi.org/10.18653/v1/p19-1656}
\BIBentrySTDinterwordspacing

\bibitem{Gao2025SpokenIJ}
\BIBentryALTinterwordspacing
X.~Gao, S.~Nayak, and M.~Coler, ``Spoken in jest, detected in earnest: A
  systematic review of sarcasm recognition—multimodal fusion, challenges, and
  future prospects,'' \emph{IEEE Transactions on Affective Computing}, vol.~16,
  pp. 2526--2544, 2025. [Online]. Available:
  \url{https://api.semanticscholar.org/CorpusID:281194741}
\BIBentrySTDinterwordspacing

\bibitem{Raghuvanshi2025IntramodalRA}
\BIBentryALTinterwordspacing
D.~Raghuvanshi, X.~Gao, Z.~Li, S.~Bansal, M.~Coler, N.~Kumar, and S.~Nayak,
  ``Intra-modal relation and emotional incongruity learning using graph
  attention networks for multimodal sarcasm detection,'' \emph{ICASSP 2025 -
  2025 IEEE International Conference on Acoustics, Speech and Signal Processing
  (ICASSP)}, pp. 1--5, 2025. [Online]. Available:
  \url{https://api.semanticscholar.org/CorpusID:276977265}
\BIBentrySTDinterwordspacing

\bibitem{li2025leveraginglargelanguagemodels}
Z.~Li, Y.~Zhang, X.~Gao, S.~Nayak, and M.~Coler, ``{Leveraging Large Language
  Models for Sarcastic Speech Annotation in Sarcasm Detection},'' in
  \emph{{Interspeech 2025}}, 2025, pp. 3973--3977.

\bibitem{scott2025musagmultimodalgermansarcasm}
\BIBentryALTinterwordspacing
A.~Scott, M.~Z{\"u}fle, and J.~Niehues, ``Musag: A multimodal german sarcasm
  dataset with full-modal annotations,'' \emph{ArXiv}, vol. abs/2510.24178,
  2025. [Online]. Available:
  \url{https://api.semanticscholar.org/CorpusID:282400655}
\BIBentrySTDinterwordspacing

\bibitem{adieu16}
G.~Trigeorgis, F.~Ringeval, R.~Brueckner, E.~Marchi, M.~A. Nicolaou,
  B.~Schuller, and S.~Zafeiriou, ``Adieu features? end-to-end speech emotion
  recognition using a deep convolutional recurrent network,'' in \emph{2016
  IEEE International Conference on Acoustics, Speech and Signal Processing
  (ICASSP)}, 2016, pp. 5200--5204.

\bibitem{lakshminarayanan2017simplescalablepredictiveuncertainty}
B.~Lakshminarayanan, A.~Pritzel, and C.~Blundell, ``Simple and scalable
  predictive uncertainty estimation using deep ensembles,'' in
  \emph{Proceedings of the 31st International Conference on Neural Information
  Processing Systems}, ser. NIPS'17.\hskip 1em plus 0.5em minus 0.4em\relax Red
  Hook, NY, USA: Curran Associates Inc., 2017, p. 6405–6416.

\bibitem{kendall}
\BIBentryALTinterwordspacing
A.~Kendall and Y.~Gal, ``What uncertainties do we need in bayesian deep
  learning for computer vision?'' in \emph{Advances in Neural Information
  Processing Systems}, I.~Guyon, U.~V. Luxburg, S.~Bengio, H.~Wallach,
  R.~Fergus, S.~Vishwanathan, and R.~Garnett, Eds., vol.~30.\hskip 1em plus
  0.5em minus 0.4em\relax Curran Associates, Inc., 2017. [Online]. Available:
  \url{https://proceedings.neurips.cc/paper_files/paper/2017/file/2650d6089a6d640c5e85b2b88265dc2b-Paper.pdf}
\BIBentrySTDinterwordspacing

\bibitem{baevski2020wav2vec20frameworkselfsupervised}
\BIBentryALTinterwordspacing
A.~Baevski, Y.~Zhou, A.~Mohamed, and M.~Auli, ``wav2vec 2.0: A framework for
  self-supervised learning of speech representations,'' in \emph{Advances in
  Neural Information Processing Systems}, H.~Larochelle, M.~Ranzato,
  R.~Hadsell, M.~Balcan, and H.~Lin, Eds., vol.~33.\hskip 1em plus 0.5em minus
  0.4em\relax Curran Associates, Inc., 2020, pp. 12\,449--12\,460. [Online].
  Available:
  \url{https://proceedings.neurips.cc/paper_files/paper/2020/file/92d1e1eb1cd6f9fba3227870bb6d7f07-Paper.pdf}
\BIBentrySTDinterwordspacing

\bibitem{HsuBTLSM21}
\BIBentryALTinterwordspacing
W.~Hsu, B.~Bolte, Y.~H. Tsai, K.~Lakhotia, R.~Salakhutdinov, and A.~Mohamed,
  ``Hubert: Self-supervised speech representation learning by masked prediction
  of hidden units,'' \emph{{IEEE} {ACM} Trans. Audio Speech Lang. Process.},
  vol.~29, pp. 3451--3460, 2021. [Online]. Available:
  \url{https://doi.org/10.1109/TASLP.2021.3122291}
\BIBentrySTDinterwordspacing

\bibitem{WavLM}
S.~Chen, C.~Wang, Z.~Chen, Y.~Wu, S.~Liu, Z.~Chen, J.~Li, N.~Kanda,
  T.~Yoshioka, X.~Xiao, J.~Wu, L.~Zhou, S.~Ren, Y.~Qian, Y.~Qian, J.~Wu,
  M.~Zeng, X.~Yu, and F.~Wei, ``Wavlm: Large-scale self-supervised pre-training
  for full stack speech processing,'' \emph{IEEE Journal of Selected Topics in
  Signal Processing}, vol.~16, no.~6, pp. 1505--1518, 2022.

\bibitem{dai21_interspeech}
X.~Dai, C.~Gong, L.~Wang, and K.~Zhang, ``Information sieve: Content leakage
  reduction in end-to-end prosody transfer for expressive speech synthesis,''
  in \emph{Interspeech 2021}, 2021, pp. 131--135.

\bibitem{crossling}
S.~Khan, I.~Qasim, W.~Khan, A.~Khan, J.~Khan, A.~Qahmash, and Y.~Ghadi, ``An
  automated approach to identify sarcasm in low-resource language,'' \emph{PLOS
  ONE}, vol.~19, 12 2024.

\bibitem{GeMAPS}
F.~Eyben, K.~R. Scherer, B.~W. Schuller, J.~Sundberg, E.~André, C.~Busso,
  L.~Y. Devillers, J.~Epps, P.~Laukka, S.~S. Narayanan, and K.~P. Truong, ``The
  geneva minimalistic acoustic parameter set (gemaps) for voice research and
  affective computing,'' \emph{IEEE Transactions on Affective Computing},
  vol.~7, no.~2, pp. 190--202, 2016.

\bibitem{pepino21_interspeech}
L.~Pepino, P.~Riera, and L.~Ferrer, ``{Emotion Recognition from Speech Using
  wav2vec 2.0 Embeddings},'' in \emph{{Interspeech 2021}}, 2021, pp.
  3400--3404.

\bibitem{electronics12030666}
\BIBentryALTinterwordspacing
A.-C. B{\u{a}}roiu and {\c{S}}.~Tr{\u{a}}u{\c{s}}an-Matu, ``Comparison of deep
  learning models for automatic detection of sarcasm context on the {MUS}t{ARD}
  dataset,'' \emph{Electronics}, vol.~12, no.~3, 2023. [Online]. Available:
  \url{https://www.mdpi.com/2079-9292/12/3/666}
\BIBentrySTDinterwordspacing

\bibitem{mused}
\BIBentryALTinterwordspacing
X.~Gao, S.~Bansal, K.~Gowda, Z.~Li, S.~Nayak, N.~Kumar, and M.~Coler, ``Amused:
  An attentive deep neural network for multimodal sarcasm detection
  incorporating bi-modal data augmentation,'' \emph{CoRR}, vol. abs/2412.10103,
  2024. [Online]. Available: \url{https://doi.org/10.48550/arXiv.2412.10103}
\BIBentrySTDinterwordspacing

\bibitem{tiwari2023}
S.~{Bhosale}, A.~{Chaudhuri}, A.~L.~R. {Williams}, D.~{Tiwari}, A.~{Dutta},
  X.~{Zhu}, P.~{Bhattacharyya}, and D.~{Kanojia}, ``{Sarcasm in Sight and
  Sound: Benchmarking and Expansion to Improve Multimodal Sarcasm Detection},''
  \emph{arXiv e-prints}, p. arXiv:2310.01430, Sep. 2023.

\end{thebibliography}

\end{document}